\makeatletter\def\graphicscache@inhibit{true}\makeatother
\documentclass[letterpaper, 10 pt, conference]{ieeeconf}  %

\IEEEoverridecommandlockouts                              %

\usepackage[utf8]{inputenc}
\usepackage{url}
\usepackage{array}
\usepackage{graphicx} %
\usepackage{subcaption}
\usepackage{siunitx}
\usepackage{amsmath} %
\usepackage{amssymb}  %
\usepackage{algorithm, algorithmic}
\usepackage{multirow}
\usepackage{diagbox}
\usepackage{tikz}
\usepackage{graphicscache}
\usepackage{tabularx}

\usepackage{caption}
\makeatletter
\let\NAT@parse\undefined
\makeatother
\usepackage{hyperref}

\newcolumntype{P}[1]{>{\centering\arraybackslash}p{#1}}
\newcommand{\etal}{et al.}
\newcommand{\figref}[1]{Fig.~\ref{#1}}
\newcommand{\tabref}[1]{Tab.~\ref{#1}}
\newcommand{\secref}[1]{Sec.~\ref{#1}}
\usepackage{xpatch}
\xpretocmd{\eqref}{Eq.~}{}{}

\title{\LARGE \bf
Directional TSDF: Modeling Surface Orientation for Coherent Meshes
}

\author{Malte Splietker and Sven Behnke%
\thanks{This research has been supported by MBZIRC 2017 price money.}%
\thanks{All authors are with the Autonomous Intelligent Systems Group, University of Bonn, Germany. {\tt\small splietke@ais.uni-bonn.de}}%
}

\usepackage{fancyhdr}
\begin{document}

\maketitle
\thispagestyle{fancy} %
\pagestyle{empty} %

\begin{abstract}

Real-time 3D reconstruction from RGB-D sensor data plays an important role in many robotic applications, such as object modeling and mapping.
The popular method of fusing depth information into a truncated signed distance function (TSDF) and applying the marching cubes algorithm for mesh extraction has severe issues with thin structures: not only does it lead to loss of accuracy, but it can generate completely wrong surfaces.
To address this, we propose the directional TSDF---a novel representation that stores opposite surfaces separate from each other. The marching cubes algorithm is modified accordingly to retrieve a coherent mesh representation. We further increase the accuracy by using surface gradient-based ray casting for fusing new measurements.
We show that our method outperforms state-of-the-art TSDF reconstruction algorithms in mesh accuracy.

\end{abstract}

\section{Introduction}
3D models are a useful way to describe objects or whole environments, which can be used in a variety of robotic applications like scene understanding, manipulation, and navigation. Since the publication of KinectFusion~\cite{Newcombe2011} in 2011, TSDF fusion has turned into a de facto standard for fast registration and reconstruction using low-cost RGB-D sensors. TSDF fusion divides the modeled volume into a discretized grid of voxels and fuses distance information into it. Even though alternative approaches based on surfel predictions~\cite{Stueckler2014, Whelan2016} or direct meshing~\cite{Greene2017, Piazza2018} have emerged, TDSF fusion still remains the most popular choice.

\begin{figure}[t]
  \centering
  \includegraphics[width=\linewidth]{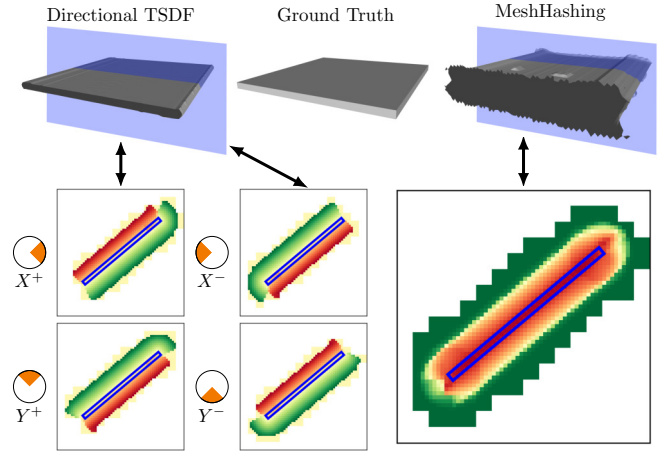}
  \caption{Directional TSDF (proposed) solves the problems of reconstructing thin objects, which state-of-the-art TSDF fusion methods have issues with (here MeshHashing~\cite{Dong2018}). The bottom row shows the TSDFs (left: directional, right: undirected) in cross section, where the blue rectangle indicates the ground truth. The green and red colors denote areas that are in front or behind the surface, respectively. Color gradients indicate the signed distances to the object and the surface is extracted at the transition between the colors.
}
  \label{fig:iconic}
\end{figure}

Since the publication of KinectFusion, much work has been done to improve reconstruction speed and quality. There are, however, some fundamental limitations within the method itself. Firstly, it cannot represent anything thinner than the voxel size. While loss of fine details would be acceptable, the extracted surfaces can become completely wrong as illustrated in \figref{fig:iconic}. Secondly, the state-of-the-art method of iteratively integrating new measurements is erroneous for steep observation angles and different observation directions as it overwrites and, in this way, destroys the representation.

\begin{figure*}[t]
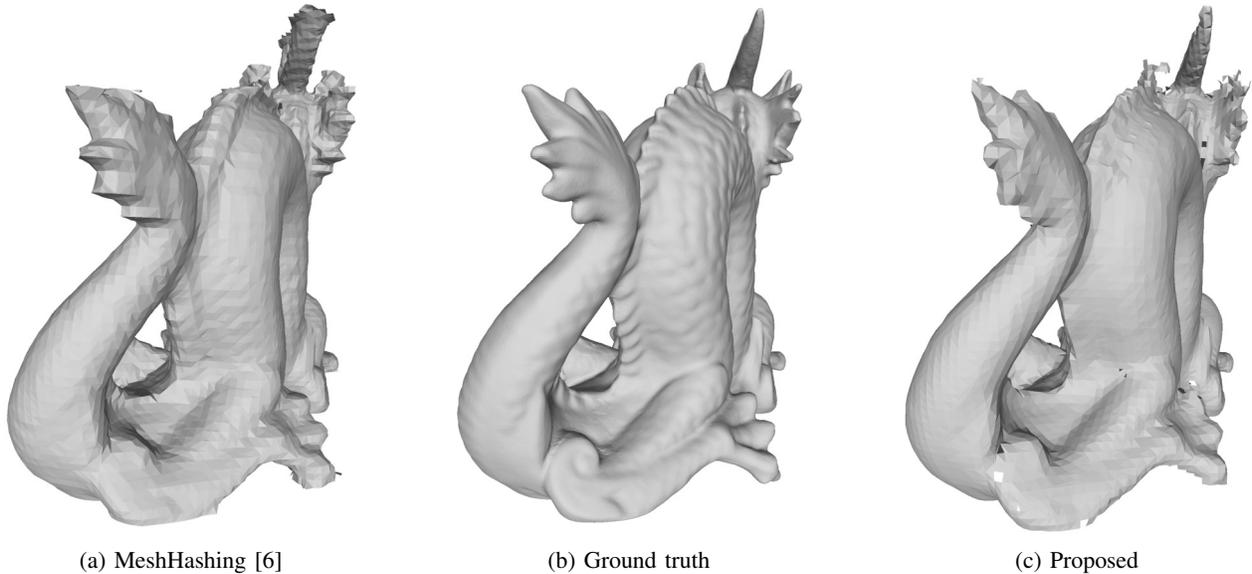

  \begin{subfigure}[b]{.33\linewidth}
    \centering
    \includegraphics[width=.8\textwidth]{images/meshes/showcase/ridge_sota.png}
    \caption{MeshHashing~\cite{Dong2018}}
  \end{subfigure}
  \begin{subfigure}[b]{.33\linewidth}
    \centering
    \includegraphics[width=.8\textwidth]{images/meshes/showcase/ridge_groundtruth.png}
    \caption{Ground truth}
  \end{subfigure}
  \begin{subfigure}[b]{.33\linewidth}
    \centering
    \includegraphics[width=.8\textwidth]{images/meshes/showcase/ridge_proposed.png}
    \caption{Proposed}
  \end{subfigure}
  \caption{Qualitative reconstruction comparison on the Dragon model from the Stanford 3D scanning repository~\cite{StanfordScanrep}. Voxel size in both cases is \SI{10}{\mm}.}
  \label{fig:qualitative_comparison}
\end{figure*}
A common way to mitigate these issues is to decrease the voxel size, which increases details and makes the problem less noticeable. However, for embedded hardware or large-scale mapping tasks a coarser voxel resolution might be required due to computation and memory restrictions. Moreover, the effect is not only affected by the voxel resolution, but also by the often depth-dependent truncation distance which is required to deal with measurement noise and usually spans multiple voxels.
The problem lies in the representation itself, as the TSDF implicitly encodes surfaces as zero crossings and the density of these transitions is bounded by the voxel resolution and the truncation distance. The direction from which surfaces have been observed is only encoded indirectly by the gradient normal. This is especially problematic during fusion, because information from different directions (at corners or on opposite sides of a wall) might be contradictory within the truncation range. This leaves the TSDF in an inconsistent state with false information. Furthermore, the state-of-the-art method for data integration, voxel projection, where each voxel is projected into the camera image and associated with the nearest pixel, has disadvantages. It causes serious aliasing, incorrectly handles steep surfaces and neglects large amounts of input data, especially for larger voxel sizes.
To address these issues, we introduce the directional TSDF---a novel data structure that encodes the surface orientations by dividing the modeled volume into six directions according to the positive and negative coordinate axes. This representation can handle observations of thin structures from different, opposing directions, without introducing aliasing. Data integration is done in a ray-casting fashion along the surface normals for every input point. The advantages are that all data is utilized and that the resulting representation is more accurate with respect to steep angle observations.
To extract surfaces from the directional TSDF, a modified marching cubes algorithm is proposed, which can also model opposite faces while remaining computationally inexpensive.

In summary, the contributions of this paper are a novel representation which is better suited for mapping scenes from different viewing directions. We also present an improved data integration scheme which considers the actual surface gradient for determining the correct voxels for fusion. Finally, a mesh extraction method for this new representation is proposed. We thoroughly evaluate our methods on standard data sets.

\section{Related Work}

Surface reconstruction from range data has been an active research topic for a long time. It gained in popularity through the availability of affordable depth cameras and parallel computing hardware. Zollhöfer \etal~\cite{Zollhoefer2018} give a comprehensive overview on modern 3D reconstruction from RGB-D data. The two main streams or research are TSDF fusion~\cite{Newcombe2011,Dong2018} and surfel extraction~\cite{Stueckler2014, Whelan2016}. TSDF-based methods make up the majority, due to their simplicity and mesh output. Surfels, however, maintain the surface and observation direction in form of a normal per surfel. Thus they can distinguish observations from different sides. Another interesting approach is presented by Sch\"ops et al.~\cite{Schoeps2018}, who triangulate surfels to create a mesh representation.

An important step in data keeping was the switch from statically allocated voxel arrays to hash tables, allocating only required areas as proposed by Niesner~\etal~\cite{Niesner2013}. This enables scanning of large areas with limited memory and is considered state-of-the-art~\cite{Dong2018, Klingensmith2015, Oleynikova2017}.
We base our work on Dong et al.~\cite{Dong2018}, who further improve the data structure by tightly coupling voxel and meshing data. Signed distance data is stored on the corners of mesh cubes, which makes interpolation superfluous. Also the allocation, storage and access of vertex information is coupled to the structure, which decreases memory consumption and computation time.

As stated earlier, a major drawback of the TSDF representation is the voxel resolution, because the maximum object resolution is proportional to the voxel size. While decreasing the voxel size is one option, it is also wasteful in many areas with little detail. Steinbrücker \etal~\cite{Steinbruecker2014} address this by dynamically adjusting the voxel resolution at the cost of additional octree nesting depth and a very complex surface extraction algorithm. This does, however, not solve the problem completely as depth-dependent noise needs to be considered in choosing the truncation range. The undirected TSDF of \figref{fig:iconic} shows, how the truncation range from the opposite direction pushes the zero crossing away from the ground truth.
Henry \etal~\cite{Henry2013} dynamically create new TSDF volumes whenever the angle of the surface changes too much. This is similar to our approach, but relies on managing a huge number of separate volumes. Also the volume separation decision relies on larger surfaces; therefore it does not deal with small details. Moreover, their approach lacks a mesh extraction method and renderings are created by ray casting.

The de facto standard method for integrating measurements, voxel projection~\cite{Newcombe2011, Dong2018, Steinbruecker2014, Henry2013}, suffers from aliasing effects especially for large voxel sizes and steep observation angles~\cite{Klingensmith2015}. %
Curless \etal~\cite{Curless1996} perform voxel projection onto an intermediate mesh generated from the input, thereby using the interpolated values of multiple input points to update a voxel.
Many approaches have tried to overcome the issues of voxel projection TSDF fusion. Commonly data integration is weighted according to the quality of measurements. Stotko and Golla~\cite{Stotko2015} evaluated different weighting options for fusion. This helps to compensate distance- and angle-dependent noise.
To reduce the effects of integrating false information from surfaces with high observation angles, the point-to-plane distance metric can be applied~\cite{Bylow2013}.
As an alternative to voxel projection, ray casting~\cite{Klingensmith2015} shoots a ray from the camera through every observed point and all voxels within the truncation range are updated. A sped up version using grouped ray casting was presented by Oleynikova et al.~\cite{Oleynikova2017}. While for larger voxels the computational overhead is higher, the advantage is that there are no aliasing effects and that all information is utilized. Fossel \etal~\cite{Fossel2015} address the issue that---especially for wide-angle sensors like LIDARs---the line-of-sight ray casting direction does not always comply with the surface direction. They estimate the surface gradient and choose the truncation range along the normal in a 2D SLAM system.

In contrast to the related works, we are proposing an improved representation based on the TSDF that utilizes the idea from Henry~\cite{Henry2013} to represent surfaces with different orientations separate from each other. The implementation is based on the work of Dong \etal~\cite{Dong2018}, which also serves as a baseline for state-of-the-art methods using voxel projection and the marching cubes algorithm. Also we apply the gradient-based ray casting concept from Fossel et al.~\cite{Fossel2015}.
The key features of our method are:
\begin{itemize}
  \item the directional TSDF representation that divides the modeled volume into six directions, thereby separately representing surfaces with different orientations,
  \item a gradient-based ray casting fusion for improved results,
  \item a thread-safe parallelization of ray casting fusion, and
  \item a modified marching cubes algorithm for mesh extraction from this representation.
\end{itemize}

\section{Directional TSDF}
A Signed Distance Function (SDF) denotes a function that for every 3D point yields the shortest distance to any surface. The sign denotes, whether the point is in front or behind the surface (inside an object).
Let $\Omega \subset \mathbb{R}^3$ be a subset of space, e.g. a number of objects. In surface reconstruction, the points of interest lie on the boundary $\partial\Omega$. For a distance function $d$ and any point $\mathbf{p} \in \mathbb{R}^3$, the SDF $\Phi$ defines the signed distance to the surface:
\begin{equation}
  \Phi:\mathbb{R}^3 \longrightarrow \mathbb{R},~\Phi(\mathbf{p}) = \left\{
 \begin{array}{ll}
   -d(\mathbf{p}, \partial\Omega) & \text{if } \mathbf{p} \in \Omega, \\
   \phantom{-}d(\mathbf{p}, \partial\Omega) & \text{if } \mathbf{p} \in \Omega^c.
\end{array}
\right.
\end{equation}
That is, points that lie inside of the object have a negative value and the surface lies exactly at the zero crossing between positive and negative values.
Consequently, most regions of the SDF are superfluous for determining the surface. The Truncated Signed Distance Function (TSDF) cuts of all values above a \emph{truncation threshold} $\tau$, so everything outside the truncation range can be omitted.
Typically TSDFs are estimated by a discretized grid of voxels and interpolation between the grid points. The truncation range is required to cope with aliasing effects and sensor noise and typically spans multiple voxels.
While the method has proven to work in many scenarios, it has limitations, especially regarding thin objects, because the voxel resolution and truncation distance dictate the minimum thickness of objects. This effect occurs when observing small structures from opposite sides, as shown on the right hand side of \figref{fig:iconic} where the object ``blows up'' as the truncation range pushes the contour further out. The cross section of the TSDF shows how far the estimated contour (transition between red and green) is away from the ground truth (blue rectangle).

The problem obviously lies in the representation itself, since a zero transition cannot be represented by fewer than two voxels (one with for each positive and negative value). We propose a new representation, called Directional TSDF (DTSDF), which stores the signed distance information in different volumes according to the surface gradient
\begin{equation}
  \Phi^d: \mathbb{R}^3 \longrightarrow \mathbb{R}^6,~\Phi^d(p) = (\Phi_D(p))_{D\in\textrm{Directions}}.
\end{equation}
The $\textrm{Directions}=\{X^+, X^-, Y^+, Y^-, Z^+, Z^-\}$ are defined by the positive and negative coordinate axes $\mathbf{v} = \{(1, 0, 0)^\intercal, (-1, 0, 0)^\intercal, \cdots \}$. This way, each voxel can encode up to six surfaces, which is beneficial for modeling orthogonal or opposite surfaces and arbitrarily thin objects.
Each direction spans a sector of applicable surface normals to determine which information belongs where. While the number of sectors can be arbitrarily high, six is an obvious choice due to the cuboid voxel shape. Fewer sectors are problematic, as the covered angle increases, though in principle four sectors spanning a tetrahedron would be sufficient.

Given a measurement point's surface normal $\mathbf{n}$ and a direction vector $\mathbf{v}_{D}$, the direction-correspondence weight is defined as
\begin{equation}
  w_{D}(\mathbf{n}) = \left< \mathbf{n}, \mathbf{v}_{D} \right>.
  \label{}
\end{equation}
A measurement is integrated into all directions, whose weight is above a threshold of $\sin(\pi/8)$. This allows for a smooth transition between the sectors, which is important for the creation of coherent meshes. Note, that each measurement point is integrated into at most three directions.

\subsection{Data Structure}
\begin{figure}[thpb]
 \centering
 \includegraphics[width=\linewidth]{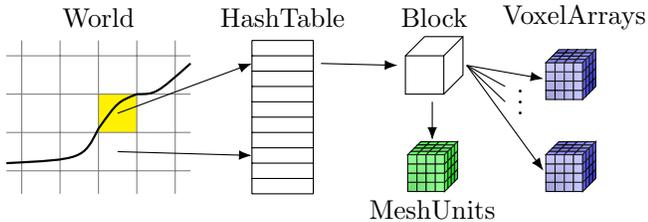}
 \caption{Data structure for dynamically allocating blocks and per-direction voxel arrays.}
 \label{fig:datastructure}
\end{figure}

Given the directional correspondence computation, it is easy to spot that not all voxels need to store information for all directions. In order to save memory, the classical voxel hashing data structure~\cite{Dong2018, Niesner2013} is extended to hold varying numbers of voxel arrays, each of which represents a certain direction. \figref{fig:datastructure} shows the connections: The block coordinates from the world are hashed. Then a conflict-resolving hash table maps to a dynamically allocated block. For every block the up to six voxel arrays are allocated as needed. For clearer visualization, the modeled volumes in all example images are depicted in 2D, but all arguments easily extend to 3D.

\subsection{Gradient-Directed Ray Casting Fusion}
\begin{figure}[thpb]
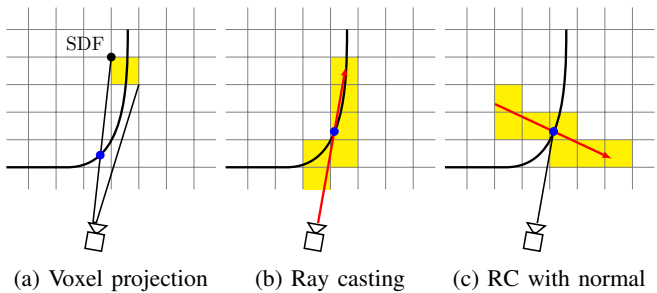

  \begin{subfigure}[b]{.325\linewidth}
     \includegraphics[width=\textwidth]{images/fusion_voxel_projection.pdf}
     \caption{Voxel projection}
     \label{fig:fusion_modes_vp}
  \end{subfigure}
  \begin{subfigure}[b]{.325\linewidth}
     \includegraphics[width=\textwidth]{images/fusion_raycasting.pdf}
     \caption{Ray casting}
     \label{fig:fusion_modes_rc}
  \end{subfigure}
  \begin{subfigure}[b]{.325\linewidth}
     \includegraphics[width=\textwidth]{images/fusion_raycasting_normal.pdf}
     \caption{RC with normal}
     \label{fig:fusion_modes_rcn}
  \end{subfigure}
  \caption{Fusion mode comparison. A measured surface point (blue dot) of the ground truth surface (black curve) is used to update voxels (yellow squares) along the fusion ray (red).}
 \label{fig:fusion_modes}
\end{figure}

Another novelty of our work is the surface gradient based ray casting fusion which, to our knowledge, has not been applied in 3D TSDF fusion before. The default method for fusing new measurements into the TSDF is voxel projection (VP), where voxels in view range and inside the camera frustum are projected onto the camera plane and are then associated with a single pixel in the depth image.
\figref{fig:fusion_modes_vp} depicts an example of the drawbacks pointed out before, where the projected SDF point (black) is far away from the projected measurement (red dot), thus gets a high SDF value, even though it is very close to a surface. Handling misassociations like those can be partially mended by using the point-to-plane metric for updating the SDF value~\cite{Bylow2013}.
\figref{fig:fusion_modes_rc} shows ray casting fusion, where the algorithm casts a ray through every depth pixel and all intersected voxels within truncation range around the surface point are updated~\cite{Klingensmith2015, Oleynikova2017}.
While improving the association problem and better utilizing the available data, steep observation angles remain problematic~\cite{Fossel2015}. Our experiments have shown that this method, especially using the standard TSDF, worsens corners when rays shoot through them from different directions.
Instead we extend the idea from Fossel \etal~\cite{Fossel2015} to our method and use the surface normals as shown in \figref{fig:fusion_modes_rcn}. Starting from the projected surface point, a ray is cast along the normal (in both directions) and all intersecting voxels in truncation range are updated. To circumvent expensive voxel-ray intersection computations, the algorithm applies voxel traversal as proposed in~\cite{Amanatides1987}.

As the SDF corners of traversed voxels stray left and right from the ray, we furthermore apply the point-to-plane metric to increase the accuracy of the SDF. For a given surface point $\mathbf{p}$ and corresponding normal $\mathbf{n}_{\mathbf{p}}$ the distance function is
\begin{equation}
  d_{\mathrm{p2pl}}(\mathbf{x}) = (\mathbf{p} - \mathbf{x})^\intercal \mathbf{n}_{\mathbf{p}}.
\end{equation}
The normals are computed using simple neighborhood estimation on the depth image. Due to noise and discretization in the depth image the estimated normals can be inaccurate, so a bilateral filter is applied. The depth image remains unfiltered to preserve reconstruction detail. While voxel projection updates every voxel exactly once, ray casting requires multiple updates per iteration. Details on the thread-safe fusion implementation are explained in \secref{sec:thread_safe_fusion}.

For the SDF update a combined weighting scheme is applied. The noise of RGB-D cameras depends on the measured distance, which is accounted for in $w_{\mathrm{depth}}$. A high surface to view direction angle also increases inaccuracy, so it is down-weighted by $w_{\mathrm{angle}}$. The factor $w_{D}$, defined above, works the same way as $w_{\mathrm{angle}}$, but down-weights measurements that do not comply with the current fusion direction $D$.
The combined weight is
\begin{align}
  w &= w_{\mathrm{depth}} \cdot w_{\mathrm{angle}} \cdot w_{\mathrm{D}}.
\end{align}
Stotko~\cite{Stotko2015} gives a detailed overview on weighting factors.

\section{Directional Marching Cubes}
A straight-forward and efficient method for mesh extraction from the TSDF representation is the marching cubes (MC) algorithm~\cite{Lorensen1987}. It can be easily parallelized. The world is again divided into a regular grid of \emph{mesh units}, which in this implementation is identical to the voxel grid. For every mesh unit, the SDF values at the corners are checked and for all edges which contain a zero transition, an appropriate set of triangles is generated. Since there are only 256 configurations for positive and negative corners, called \emph{MC index}, a lookup table is used for efficiency.

\begin{figure}[thpb]
 \centering
 \includegraphics[width=\linewidth]{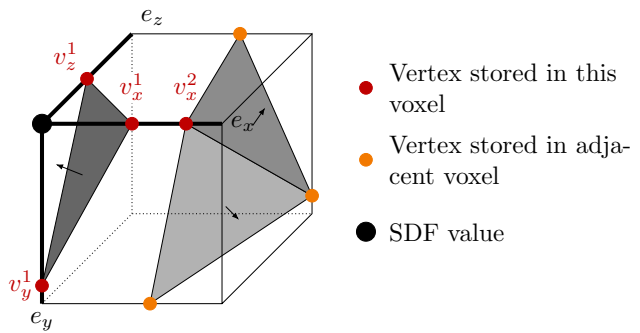}
 \caption{Every mesh unit is responsible for storing vertices (red dots) on the thick edges $e_x, e_y, e_z$. If triangles have vertices on other edges, these are stored in adjacent voxels (yellow dots).}
 \label{fig:mc_datastructure}
\end{figure}

For the directional TSDF, finding those zero transitions becomes more complicated as it is not immediately clear how to combine the information collected for different directions. Also there can be opposite faces within the same mesh unit, which the original algorithm cannot handle. Every edge must now be able to hold up to one zero transition per direction. Therefore, the data structure from~\cite{Dong2018} is extended as depicted in \figref{fig:mc_datastructure}. The data structure makes SDF interpolation superfluous as the relevant SDF values can be directly fetched.
Every mesh unit handles three edges, each of which can have up to two vertices for opposite surfaces. Triangles may have  vertices on other edges, which are stored in adjacent mesh units.
The outline of directional marching cubes is described in Algorithm~\ref{alg:directional_mc}. The steps are explained throughout this section.
\begin{algorithm}
    \caption{MC Index Combining}
    \label{alg:directional_mc}
    \begin{algorithmic}[1]
      \FOR {every mesh unit}
        \FOR {every direction $D$}
            \STATE get MC index $\mathrm{mc}_D$, SDF weights $w_D$
        \ENDFOR
        \STATE Directional MC index filtering
        \STATE Inter-directional filtering
        \STATE Compute surface offsets for each edge
        \STATE Determine combined MC indices (up to 2)
        \STATE Allocate Vertices and Edges
    \ENDFOR
    \end{algorithmic} 
\end{algorithm}

\subsection{Filtering}
It is necessary to filter the MC indices, as due to the nature of the representation some degree of false information occurs. Especially at the edge of the TSDF or at spots where different geometry collides, deviation and overhangs appear.
Filtering is done in two stages, intra-directional and inter-directional.

\begin{figure}[thpb]
 \centering
 \includegraphics{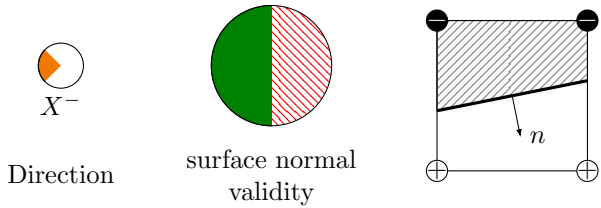}
 \caption{Filtering an implausible MC index by surface normal. The normal $\mathbf{n}$ exceeds the validity range (green semicircle) for direction $X^-$. The white and black corners of the mesh unit indicate whether the value is in front of or behind the surface, respectively. }
 \label{fig:filter_direction}
\end{figure}
Firstly, all surfaces that would contradict the respective directions are discarded. This is performed using a pre-computed lookup table and comparison in cases where the orientation of the surface decides. \figref{fig:filter_direction} shows an example, where the normal of the potential surface is outside the valid range for direction $X^-$, so it is discarded. In this case the simple table lookup is insufficient, because there are configurations with the same MC index, where the normal is inside the valid range. The SDF gradient is utilized to identify these outliers.

\begin{figure}[thpb]
 \includegraphics[width=\linewidth]{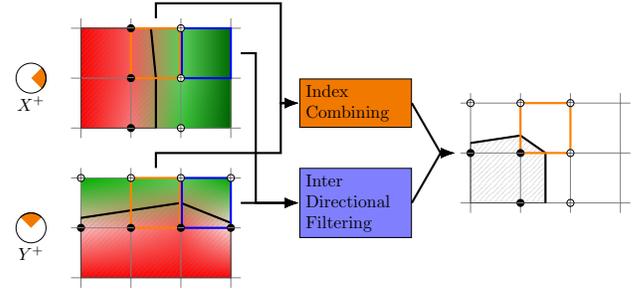}
 \caption{MC index combining (orange) and inter-directional filtering (blue) is applied to TSDFs of directions $X^+$ and $Y^-$ to retrieve the combined surface on the right side.}
 \label{fig:inter_directional_combining}
\end{figure}

In the second step, the different directions are weighted against each other in order to decide, whether a hypothesized surface is correct. The blue mesh unit in \figref{fig:inter_directional_combining} shows a typical example, how the overhanging edge of direction $Y^+$ is identified as a false positive by $X^+$. To reduce false cancellations, the decision is made using a voting scheme. The credibility of a direction's information is accounted for by the SDF weight and surface gradient w.r.t. this direction.
For SDF weight $w^{\mathrm{sdf}}_D$, voxel center gradient $\nabla \Phi_D$, direction vector $\mathbf{v}_D$ and vote $a_D$,
\begin{equation}
  \label{eq:voting}
  a = \sum_{D} w^{\mathrm{sdf}}_D \left<\nabla \Phi_D , \mathbf{v}_D \right> a_D,\quad a_D \in \{-1, 1\}
\end{equation}
yields the consensus. If $a$ is smaller than 0, the MC index is set to 0 and no surface is extracted.

\subsection{Surface Offset Estimation}
From the remaining directions, the combined vertex positions are computed similar to what is described in~\cite{Dong2018}.
For every edge the MC indices of all directions are checked for potential zero transitions. The offset is added to a weighted average with the same combined weight used in \eqref{eq:voting}.
Since there can be opposite surfaces sharing the same edge, two offsets are stored per edge and the data structure is updated accordingly (c.f. \figref{fig:mc_datastructure}).

\subsection{Combined MC Index}
After filtering false positives there might still be a mesh unit with multiple surface hypotheses from different directions. There are multiple ways for combining these, but intersection has shown good results. The orange encircled voxels in \figref{fig:inter_directional_combining} are combined, such that a connection between the surfaces of the two directions is established.
This combination is computationally cheap, as it can be done entirely on level of MC indices. Algorithm~\ref{alg:mc_index_combining} shows how the index-wise intersection is performed. The MC index is split into it's up to four unconnected components. For each of these components the compatibility to the already combined indices is checked. The intersection of the indices is equivalent to the binary \emph{and} operation.
\begin{algorithm}
    \caption{MC Index Combining}
    \label{alg:mc_index_combining}
    \begin{algorithmic}[1]
    \renewcommand{\algorithmicrequire}{\textbf{Input:}}
    \renewcommand{\algorithmicensure}{\textbf{Output:}}
    \REQUIRE MCIndex[6]
    \ENSURE combined[2] 
    \STATE combined = (0, 0)
    \FOR {$D = 0$ to $5$}
        \FOR {component in MCIndex[D]}
            \IF {compatible(component, combined[0])}
                \STATE combined[0] \&= component
                \STATE \textbf{break}
            \ELSIF {compatible(component, combined[1])}
                \STATE combined[1] \&= component
            \ENDIF
        \ENDFOR
    \ENDFOR
    \RETURN combined
    \end{algorithmic} 
\end{algorithm}

\begin{figure}[thpb]
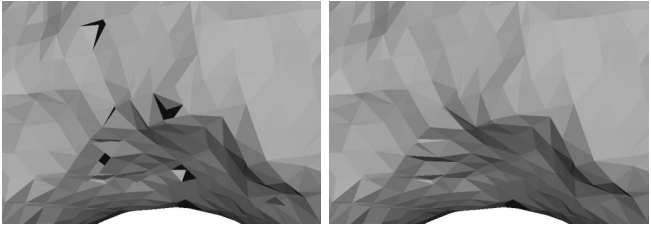

  \begin{subfigure}[b]{.49\linewidth}
     \includegraphics[width=\textwidth]{images/regularization_before.png}
     \caption{before}
  \end{subfigure}
  \begin{subfigure}[b]{.49\linewidth}
     \includegraphics[width=\textwidth]{images/regularization_after.png}
     \caption{after}
  \end{subfigure}
 \caption{Voxel neighborhood MC index regularization.}
 \label{fig:regularization}
\end{figure}

Before extracting the final mesh, regularization between MC indices of neighboring voxels is performed for all modified blocks. This helps to reduce slits and overhangs induced by previous steps and close the surface. \figref{fig:regularization} shows a mesh before and after the regularization. The procedure works solely on the MC indices by minimizing irregularities across voxel borders.

\section{Thread-Safe Ray Casting Fusion}
\label{sec:thread_safe_fusion}

The integration of new measurements into the TSDF is done by a weighted cumulative moving average. Let $D_t$ and $W_t$ be a voxel's signed distance and weight values as time $t$. $d_t$ and $w_t$ are signed distance and weight update factors which are to be integrated. Then the update is
\begin{align}
 \label{eq:wcma1}
 D_t &= \frac{W_{t-1} D_{t-1} + w_t d_t}{W_{t-1} + w_{t}} = \frac{\sum_{i=1}^t w_i d_i}{\sum_{i=1}^t w_i},\\
 W_t &= W_{t-1} + w_t.
 \label{eq:wcma2}
\end{align}
In contrast to voxel-projection fusion, ray casting leads to multiple SDF updates per voxel within the same iteration. While being mathematically sound, it is problematic for a parallel implementation, as \eqref{eq:wcma1} and \eqref{eq:wcma2} cannot be performed atomically by most hardware. \eqref{eq:wcma1} implies, that instead of applying an incremental update step for every pixel affecting a voxel, the following equivalent and thread-safe operation can be performed.
Let $S_d, S_w$ be per-voxel summation values for signed distance and weight, respectively. They are initialized with zero at the beginning of each update iteration.
\begin{align}
 S_d & \overset{atomic}{+=} w_i d_i, &S_w \overset{atomic}{+=} w_i
\end{align}
In the second step all modified voxels are iterated and the final update is computed as follows:
\begin{align}
 D_t &= \frac{W_{t-1} D_{t-1} + S_d}{W_{t-1} + S_w},\qquad W_t = W_{t-1} + S_w.
\end{align}

\section{Evaluation}
\begin{figure*}[t]
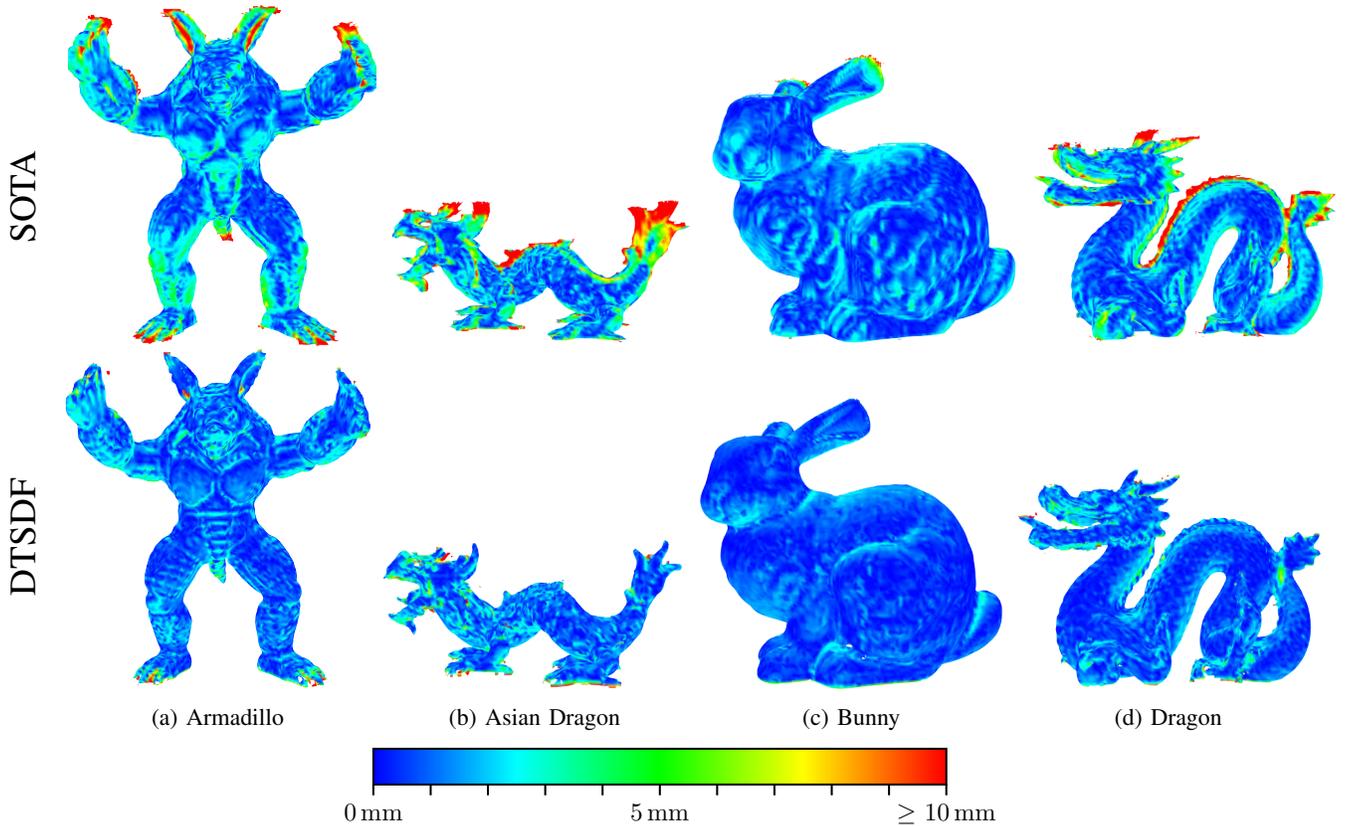

  \begin{subfigure}[b]{.04\textwidth}
    \begin{tikzpicture}
      \node[rotate=90, minimum width=4cm] at (0, 0) {\Large{SOTA}};
    \end{tikzpicture}
  \end{subfigure}
  \begin{subfigure}[b]{.23\textwidth}
    \includegraphics[width=\textwidth]{images/meshes/heatmap/armadillo_sota_10_cropped.png}
  \end{subfigure}
  \begin{subfigure}[b]{.23\textwidth}
     \includegraphics[width=\textwidth]{images/meshes/heatmap/asian_dragon_sota_10_cropped.png}
  \end{subfigure}
  \begin{subfigure}[b]{.23\textwidth}
     \includegraphics[width=\textwidth]{images/meshes/heatmap/bunny_sota_10_cropped.png}
  \end{subfigure}
  \begin{subfigure}[b]{.23\textwidth}
     \includegraphics[width=\textwidth]{images/meshes/heatmap/dragon_sota_10_cropped.png}
  \end{subfigure}
  \begin{subfigure}[b]{.04\textwidth}
    \begin{tikzpicture}
      \node[rotate=90, minimum width=5cm] at (0, 0) {\Large{DTSDF}};
    \end{tikzpicture}
  \end{subfigure}
  \begin{subfigure}[b]{.23\textwidth}
    \includegraphics[width=\textwidth]{images/meshes/heatmap/armadillo_proposed_10_cropped.png}
    \caption{Armadillo}
  \end{subfigure}
  \begin{subfigure}[b]{.23\textwidth}
     \includegraphics[width=\textwidth]{images/meshes/heatmap/asian_dragon_proposed_10_cropped.png}
    \caption{Asian Dragon}
  \end{subfigure}
  \begin{subfigure}[b]{.23\textwidth}
     \includegraphics[width=\textwidth]{images/meshes/heatmap/bunny_proposed_10_cropped.png}
    \caption{Bunny}
  \end{subfigure}
  \begin{subfigure}[b]{.23\textwidth}
     \includegraphics[width=\textwidth]{images/meshes/heatmap/dragon_proposed_10_cropped.png}
    \caption{Dragon}
  \end{subfigure}
  \center
  \begin{subfigure}[b]{.5\textwidth}
     \includegraphics[width=\textwidth]{images/heatmap.pdf}
  \end{subfigure}
  \caption{Distance error heatmap comparison between state-of-the-art (top) and proposed (bottom) on the Stanford dataset. Voxel size is \SI{10}{\mm}.}
  \label{fig:heatmap_comparison}
\end{figure*}

MeshHashing~\cite{Dong2018} serves as a baseline for recent TSDF RGB-D reconstruction algorithms with voxel hashing and marching cubes. Here, it is sometimes denoted as state-of-the-art (SOTA). Our proposed method is referred to as DTSDF. Registration is currently not implemented, so we compare the reconstruction quality and computation time.

For comparability, the well-known datasets by Zhou \etal~\cite{Zhou2013} and the Stanford Computer Graphics Laboratory~\cite{StanfordScanrep} are used. The Zhou dataset already provides trajectory and scans. For the Stanford dataset the 3D models are scaled, such that the longest side equals one meter. The camera performs an even circular motion with a two meter radius around the object, acquiring 1000 depth images with ground truth poses. The renderer uses the standard Kinect model $(f, c_x, c_y) = (525, 319.5, 239.5)$ with resolution $640\times480$ pixels. All experiments were run on a notebook with an Intel i7-4710HQ CPU (2.50GHz) and a GTX960M.

The truncation distance is fixed to four times the voxel size, as suggested by Oleynikova et al.~\cite{Oleynikova2017}. Smaller factors tend to create holes in some spots, especially for voxel projection. No other parameters were changed.

\figref{fig:qualitative_comparison} shows a qualitative comparison between SOTA, DTSDF and the ground truth at \SI{10}{\mm} voxel size. While at the given voxel resolution the amount of detail is limited, our method maintains a better surface (tail) and preserves geometry thinner than the voxel size (ridge, tail tip). The SOTA tends to enlarge the overall object.
To quantify these findings we conducted a number of experiments at different voxel resolutions and measured the RMSE against the ground truth model. \tabref{tab:rmse_voxel_size} and \tabref{tab:rmse_voxel_size_zhou} show the findings for different resolutions. Most notably, our method outperforms the SOTA in almost all cases by a good margin. Note, that we chose larger voxel sizes for the Zhou dataset due to the larger overall size of the scenes.
\begin{table}[ht]
\setlength\abovecaptionskip{-2mm}
  \caption{RMSE (in \SI{}{\mm}) of state-of-the-art and proposed method under different voxel sizes on the Stanford dataset.}
\label{tab:rmse_voxel_size}
 \begin{center}
   \begin{tabular}{ll|cccccc}
     & & \multicolumn{6}{c}{Voxel Size [\SI{}{\mm}]} \\
     dataset & mode & 5 & 10 & 20 & 30 & 40 & 50 \\
     \hline
     \multirow{2}{*}{\shortstack[l]{Arma-\\dillo}} & SOTA &
1.55 & 3.67 & 11.83 & 25.27 & 41.02 & 58.36 \\ 
     & DTSDF &
1.14 & 1.74 & 3.58 & 6.53 & 14.56 & 21.39 \\ 
     \hline
     \multirow{2}{*}{\shortstack[l]{Asian\\Dragon}} & SOTA &
2.34 & 7.11 & 19.94 & 38.70 & 55.17 & 67.86 \\
     & DTSDF &
1.56 & 2.87 & 6.81 & 11.90 & 19.00 & 30.74 \\
     \hline
     \multirow{2}{*}{\shortstack[l]{Bunny}} & SOTA &
1.90 & 3.82 & 9.01 & 17.70 & 27.97 & 37.98 \\
     & DTSDF &
0.98 & 1.23 & 2.18 & 4.17 & 8.59 & 18.37 \\
     \hline
     \multirow{2}{*}{\shortstack[l]{Dragon}} & SOTA &
1.73 & 4.44 & 12.09 & 22.40 & 35.93 & 52.67 \\ 
     & DTSDF &
1.23 & 2.15 & 5.35 & 8.99 & 11.70 & 18.70
  \end{tabular}
 \end{center}
\end{table}

\begin{table}[ht]
\setlength\abovecaptionskip{-2mm}
\setlength\belowcaptionskip{-5.5mm}
  \caption{RMSE (in \SI{}{\mm}) of state-of-the-art and proposed method under different voxel sizes on the Zhou dataset.}
\label{tab:rmse_voxel_size_zhou}
 \begin{center}
   \begin{tabular}{ll|cccc}
     & & \multicolumn{4}{c}{Voxel Size [\SI{}{\mm}]} \\
     dataset & mode & 25 & 50 & 75 & 100\\
     \hline
     \multirow{2}{*}{\shortstack[l]{Burghers}} & SOTA &
12.80 & 27.28 & 57.69 & 102.46 \\
     & DTSDF &
6.86 & 13.38 & 19.60 & 26.53 \\ 
     \hline
     \multirow{2}{*}{\shortstack[l]{Cactus-\\garden}} & SOTA &
12.78 & 32.40 & 59.35 & 85.96 \\
     & DTSDF &
13.19 & 16.69 & 33.00 & 41.68 \\
     \hline
     \multirow{2}{*}{\shortstack[l]{Lounge}} & SOTA &
14.96 & 33.91 & 57.85 & 85.52 \\
     & DTSDF &
12.41 & 25.11 & 41.95 & 51.86 \\
     \hline
     \multirow{2}{*}{\shortstack[l]{Copyroom}} & SOTA &
23.17 & 26.25 & 30.66 & 39.82 \\
     & DTSDF &
14.08 & 27.89 & 36.62 & 45.90 \\
     \hline
     \multirow{2}{*}{\shortstack[l]{Stonewall}} & SOTA &
7.23 & 13.84 & 26.22 & 39.71 \\
     & DTSDF &
6.89 & 11.47 & 18.85 & 31.20 \\
     \hline
     \multirow{2}{*}{\shortstack[l]{Totempole}} & SOTA &
10.41 & 15.19 & 27.65 & 43.00 \\
     & DTSDF &
7.86 & 9.80 & 14.88 & 24.93
  \end{tabular}
 \end{center}
\end{table}
\figref{fig:heatmap_comparison} shows a distance error heatmap visualization which matches this observation. The models reconstructed by the SOTA have hotspots, wherever there is thin geometry (ears, tail, and ridge). But other areas benefit from the DTSDF, as well.

To examine the impact of the surface gradient ray casting fusion we conducted another experiment. \tabref{tab:rmse_mode} shows the RMSE at \SI{10}{\mm} voxel resolution for different algorithm modes, which are encoded as follows: \emph{Def} and \emph{Dir} (TSDF or DTSDF), voxel projection (VP), point-to-plane (P2PL), ray casting (RC) and ray casting along normals (RCN).
\begin{table}[ht]
  \caption{RMSE (in \SI{}{\mm}) for different fusion modes, voxel size = \SI{10}{mm}.}
\label{tab:rmse_mode}
 \begin{center}
   \begin{tabularx}{\linewidth}{l|XXXXXX}
     \diagbox{model}{mode} &
     \shortstack[l]{Def\\+VP} &
     \shortstack[l]{Def\\+RC\\+P2PL} &
     \shortstack[l]{Def\\+RCN\\+P2PL} &
     \shortstack[l]{Dir\\+VP} &
     \shortstack[l]{Dir\\+RC\\+P2PL} &
     \shortstack[l]{Dir\\+RCN\\+P2PL} \\
     \hline
Armadillo & 3.670 & 4.915 & 4.839 & 1.931 & 2.478 & \textbf{1.741} \\ 
Asian Dragon & 7.106 & 10.148 & 9.211 & 3.016 & 3.545 & \textbf{2.865} \\
Bunny & 3.816 & 2.792 & 2.958 & 1.625 & 1.674 & \textbf{1.229} \\ 
Dragon & 4.438 & 6.570 & 6.169 & 2.534 & 2.930 & \textbf{2.146}
  \end{tabularx}
 \end{center}
\end{table}
Using ray casting on the default TSDF in most cases actually worsens the results, except for the bunny. This matches our earlier observation, because standard ray casting causes overshooting at corners while ray casting along normals inflates the thin parts to the maximum extent of the truncation distance. The DTSDF outperforms the SOTA even using voxel projection, but adding ray casting along normals further widens this margin. Standard ray casting, however, leads to the same problems. Only the bunny model benefits from it, because of its round shape and lack of thin, sharp edges.

\begin{figure}[thpb]
 \includegraphics[width=\linewidth]{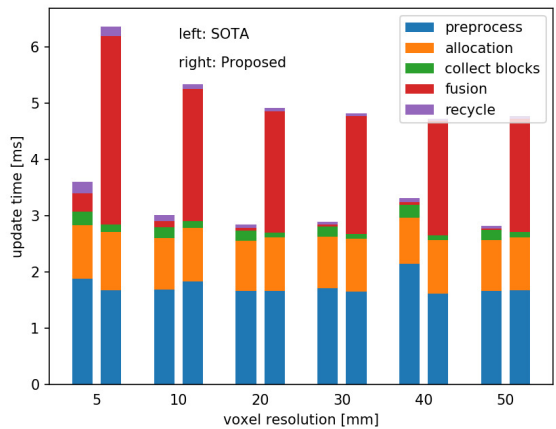}
 \caption{Mean data integration time for dataset Asian Dragon and different voxel resolutions. The left bars show the SOTA, the right side corresponds to DTSDF.}
 \label{fig:mean_time}
\end{figure}

The runtime of the algorithm can be split into two main components: data integration and meshing. \tabref{tab:total_time} shows the total update time and proportion used for meshing on the Asian Dragon dataset. As expected, the amount of time spent on meshing increases with smaller voxel sizes. The data integration, however, remains fast even for higher resolutions: \figref{fig:mean_time} breaks down the integration step for SOTA and our proposed method. Preprocessing and allocating blocks and voxel arrays make up a significant constant factor. Fusion and recycling times are affected by the number of blocks in the view frustum. The ray casting has an additional (constant) proportion determined by the number of depth pixels, but also increases with the number of voxels due to the update accumulation step (c.f. \secref{sec:thread_safe_fusion}).

\begin{table}[ht]
  \caption{Mean total update time (in ms) and percentage that is used for meshing for dataset Asian Dragon.}
  \label{tab:total_time}
 \begin{center}
   \begin{tabular}{ll|cccccc}
     & & \multicolumn{6}{c}{Voxel Size [\SI{}{\mm}]} \\
     & mode & 5 & 10 & 20 & 30 & 40 & 50 \\
     \hline
     \multirow{2}{*}{\shortstack[l]{total\\time}} & SOTA & 112.4 & 30.9 & 10.7 & 7.5 & 6.4 & 5.2 \\ 
     & Proposed & 199.0 & 74.0 & 31.7 & 19.9 & 15.1 & 13.0 \\
     \hline
     \multirow{2}{*}{\shortstack[l]{total\\time}} & SOTA & 96.8 & 90.3 & 73.5 & 61.5 & 48.0 & 45.5 \\ 
     & Proposed & 96.8 & 92.8 & 84.5 & 75.9 & 68.8 & 63.5
  \end{tabular}
 \end{center}
\end{table}

\begin{figure}[thpb]
  \includegraphics[width=\linewidth]{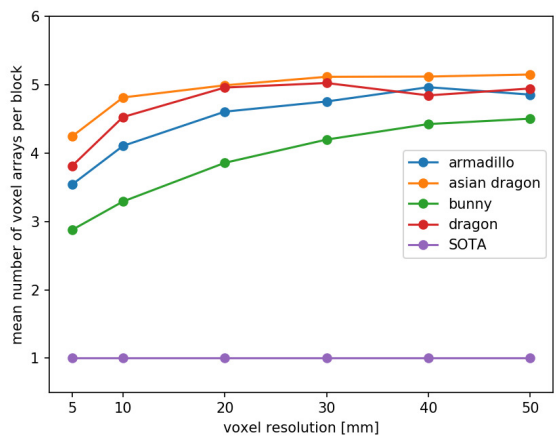}
 \caption{Mean number of voxel arrays per block for different datasets and voxel resolutions.}
 \label{fig:mean_num_voxel_arrays}
\end{figure}

Dong \etal~\cite{Dong2018} provide an accurate memory analysis for their MeshHashing implementation. The memory requirements for our algorithm differ from the original algorithm only by the number of voxel arrays allocated per block. \figref{fig:mean_num_voxel_arrays} shows the mean number of voxel arrays per block for different voxel resolutions and models. With smaller voxels the number of allocations decreases, as the number of integrations from opposite directions decreases within individual blocks. For the same reason the bunny, which is round and thicker in volume, requires fewer voxel arrays.

\section{Conclusions}

The proposed Directional TSDF representation and the matching modified marching cubes algorithm overcome limitations of the state-of-the-art method at the price of memory consumption and computation time. Less resolution is required for dealing with corners and thin objects which makes our method interesting for large-scale applications. Its robustness against different observation angles makes DTSDF also attractive for frame-to-model registration.

\def\bibfont{\footnotesize}
\bibliographystyle{IEEEtran} %
\bibliography{references}
 
\end{document}